\documentclass{article}

\usepackage[preprint]{neurips_2024}

\usepackage[utf8]{inputenc} % allow utf-8 input
\usepackage[T1]{fontenc}    % use 8-bit T1 fonts
\usepackage{hyperref}       % hyperlinks
\usepackage{url}            % simple URL typesetting
\usepackage{booktabs}       % professional-quality tables
\usepackage{amsfonts}       % blackboard math symbols
\usepackage{nicefrac}       % compact symbols for 1/2, etc.
\usepackage{microtype}      % microtypography
\usepackage{xcolor}         % colors
\usepackage{graphicx}
\usepackage{amsmath}
\usepackage{amsthm}

\title{LLM-based Knowledge Pruning for Time Series Data Analytics on Edge-computing Devices}

\author{%
  Ruibing Jin \\
   \And
   Qing Xu \\
   \AND
  Min Wu \\
   \And
  Yuecong Xu \\
   \And
  Dan Li \\
  \And
  Xiaoli Li \\
  \And
  Zhenghua Chen
}

\begin{document}

\maketitle

\begin{abstract}
	Limited by the scale and diversity of time series data, the neural networks trained on time series data often overfit and show unsatisfacotry performances.
	In comparison, large language models (LLMs) recently exhibit impressive generalization in diverse fields. Although massive LLM based approaches are proposed for time series tasks, these methods require to load the whole LLM in both training and reference. 
	This high computational demands limit practical applications in resource-constrained settings, like edge-computing and IoT devices.
	To address this issue, we propose Knowledge Pruning (KP), a novel paradigm for time series learning in this paper. For a specific downstream task, we argue that the world knowledge learned by LLMs is much \textit{redundant} and only the related knowledge termed as "\textbf{pertinent knowledge}" is useful. Unlike other methods, our KP targets to prune the \textit{redundant} knowledge and only distill the pertinent knowledge into the target model. This reduces model size and computational costs significantly. Additionally, different from existing LLM based approaches, our KP does not require to load the LLM in the process of training and testing, further easing computational burdens. With our proposed KP, a lightweight network can effectively learn the pertinent knowledge, achieving satisfactory performances with a low computation cost. To verify the effectiveness of our KP, two fundamental tasks on edge-computing devices are investigated in our experiments, where eight diverse environments or benchmarks with different networks are used to verify the generalization of our KP. 
Through experiments, our KP demonstrates effective learning of pertinent knowledge, achieving notable performance improvements in regression (19.7\% on average) and classification (up to 13.7\%) tasks, showcasing state-of-the-art results. 
\end{abstract}

\section{Introduction}
\label{sec:intro}

With the advancement of deep learning, massive methods are proposed for time series learning across different fields such as healthcare \cite{zhao2019deep,chen2018wifi}, transportation \cite{jin2023spatio}, energy \cite{zhu2023eforecaster} and industry \cite{chen2020machine}. 
Although these approaches show significant improvements on some benchmarks, it is still challenging to generalize these methods to complex scenarios \cite{jin2023large}. 

The main issue which limits the generalization of existing time series approaches, is that different measurements are applied in the process of  time series data collection. Unlike computer vision and language, it is difficult to combine these time series datasets collected from different measurements into a large scale dataset. Limited by the scale and diversity of a single time series dataset, the generalization of trained neural network on time series data is not satisfacotry.

\begin{figure}[htbp]
	\begin{center}
		\includegraphics[width=.7\linewidth]{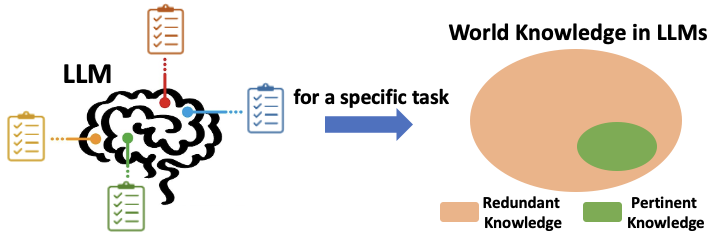}
	\end{center}
	\caption{The world knowledge learned by LLMs. Different parts of the knowledge of LLMs may contribute differently on diverse tasks. For a specific task, the world knowledge in LLMs is actually redundant and only the related knowledge termed as pertinent knowledge is useful. Our proposed \textbf{K}nowledge \textbf{P}runing (KP) aims to prune the redundant knowledge and effectively transfer the pertinent knowledge to the target model, significantly reducing computation cost while retaining satisfactory performances.}
	\label{fig:diagram}
\end{figure}

Recently, large language models (LLM) with tens of billions of parameters, demonstrate remarkable generalization capabilities in different tasks \cite{touvron2023llama,peng2023instruction}. Pre-trained on massive corpus of self-supervised data, these foundation models implicitly capture knowledge understanding on the world, which enables them to be zero-shot transferable on downstream tasks. To alleviate the issues in time series learning, some methods \cite{xue2023promptcast,chang2023llm4ts,zhou2023one,gruver2023large} are proposed to integrate the knowledge from LLMs into their frameworks.
Nevertheless,  there are two issues in these LLM based time series methods.

\begin{itemize}
	\item These approaches often require to load the whole LLM during training and inference, which is computationally expensive and time-consuming.
	\item These methods are generally based on a pre-trained and fixed LLM, which largely limits the flexible of these methods. 
\end{itemize}
Limited by these issues above, It is  challenging for existing LLM based methods to flexibly design models with different scales according to the requirements of tasks, especially for some computation constrained scenarios.

To address this problem, we re-evaluate the impact of the world knowledge acquired by LLMs on downstream tasks. We argue that \textit{for a specific downstream task, it is not necessary to transfer th entire knowledge of a pre-trained LLM into a target model.} Instead, as illustrated in Fig. \ref{fig:diagram}, we contend that this world knowledge actually can be divided into two parts: related knowledge and redundant knowledge for a specific task. Only the related knowledge termed as "\textbf{pertinent knowledge}" is what we need to transfer to the target model. Motivated by this discernment, we propose a novel compression paradigm called \textbf{K}nowledge \textbf{P}runing (KP) for LLMs, which is able to identify the pertinent knowledge, prune the redundant knowledge and effectively distill the pertinent knowledge to our target model.

Knowledge is  implicitly stored in a neural network. It is generally difficult to directly obtain a specified part of knowledge from a network. However, unlike traditional networks, LLMs can produce related knowledge description via prompts based on a dialogue scheme. According to this scheme, our proposed KP firstly generates a knowledge prompt set (KPS) for a specific task, where these prompts are forwarded to a pre-trained LLM to produce corresponding embeddings. In our proposed KP, these embeddings are called knowledge anchor points (KAPs). Although the latent space of the pertinent knowledge is a continuous space, these KAPs can be used to roughly represent this latent space. After that, a metric learning is applied to learn this  prior knowledge via knowledge distillation and transfer this pertinent knowledge to our target model. Additionally, the regression task requires a network to learn the continuous domain of the task and predict arbitrary value. To fulfill this requirement, an anchor voting scheme (AVS) is proposed, where the confidence distribution among different anchor points is generated to predict the expected output.

To verify the effectiveness of our proposed KP, massive experiments are conducted on two fundamental tasks on edge-computing devices, where different network architectures are investigated on two different task categories: classification and regression in time series learning. In classification, we evaluate the performances of our KP on four different benchmarks of human activity recognition, where our approach effectively improve the performances by up to 13.7\%. In regression, we investigate the performance of our KP on the remaining useful life prediction under four different scenarios. Through experiments, our proposed KP significantly improves the accuracy by 19.7\% on average. Our proposed KP achieves state-of-the-art performances on both tasks. Overall, our contributions are summarized as below:
\begin{itemize}
	\item  We discover that the knowledge in LLMs is much redundant for a specific downstream task. In stead of the entire knowledge, only the pertinent knowledge needs to be transferred to the target model.
	\item A novel compression paradigm, \textbf{K}nowledge \textbf{P}runing (KP) is proposed to effectively distill the pertinent knowledge into the target model, which achieves satisfactory performances, while remaining low computation cost. 
	\item An anchor voting scheme (AVS) is proposed based on the scores of knowledge anchor points to predict arbitrary value for the regression task.
	\item Experiments are extensively conducted on two fundamental tasks: classification and regression in time series learning, where different networks are employed among 8 different scenarios or benchmarks. For the regression task, our KP significantly improves the accuracy by 19.7\% on average. For the classification task, the performances are largely improved by up to 13.7\%. With our KP, state-of-the-art performances are achieved on both two tasks
\end{itemize}

\section{Related Work}
\label{sec:related}
Large language models (LLMs) recently witness significant progress and show impressive performances among a multitude of fields including natural language processing (NLP) \cite{zhao2023survey} and computer vision (CV) \cite{awais2023foundational}. To integrate the knowledge representations of LLMs into time series analytics, many approaches are proposed. PromptCast \cite{xue2023promptcast} firstly attempt to utilize LLMs for time series forecasting, where the time series data is converted into prompts. OFA  \cite{zhou2023one} proposes to fine-tune a pre-trained LLM for downstream tasks in time series analytics. Time-LLM \cite{jin2023time} and LLM4TS \cite{chang2023llm4ts} aim to repurpose a pre-trained LLM by aligning the time series domain to that of language for time series tasks. TEST \cite{sun2023test} combines the 
text prompts with time series encoding for better aligning time series data to the language. To fully utilize the generalization capability of LLMs, TEMPO \cite{cao2023tempo} augment the raw time series data by data decomposition and fine-tune a Pre-trained LLM on these augmented time series data. Although these LLM based approaches achieve significant performances on time series tasks, they are proposed based on a pre-trained and fixed LLM and require to load the whole LLM during training and inference. These drawbacks limit their flexibility and their applications on some scenarios with limited computation resources. To address this issue, we propose the knowledge pruning (KP), which is able to prune the redundant knowledge and effectively transfer the pertinent knowledge to a target model without the retaining process of LLMs, significantly reducing the computation cost and maintaining satisfactory performances.

\section{Main Work}
\label{sec:main}
Large language models (LLMs) have high computational demands and the knowledge store in them is much redundant for a specific task. To alleviate these drawbacks and facilitate the application of LLMs on computational constrained scenarios, we propose a new compression paradigm, \textbf{K}nowledge \textbf{P}runing (KP) in this section.

\begin{figure*}[t!]
	\begin{center}
		\includegraphics[width=\linewidth]{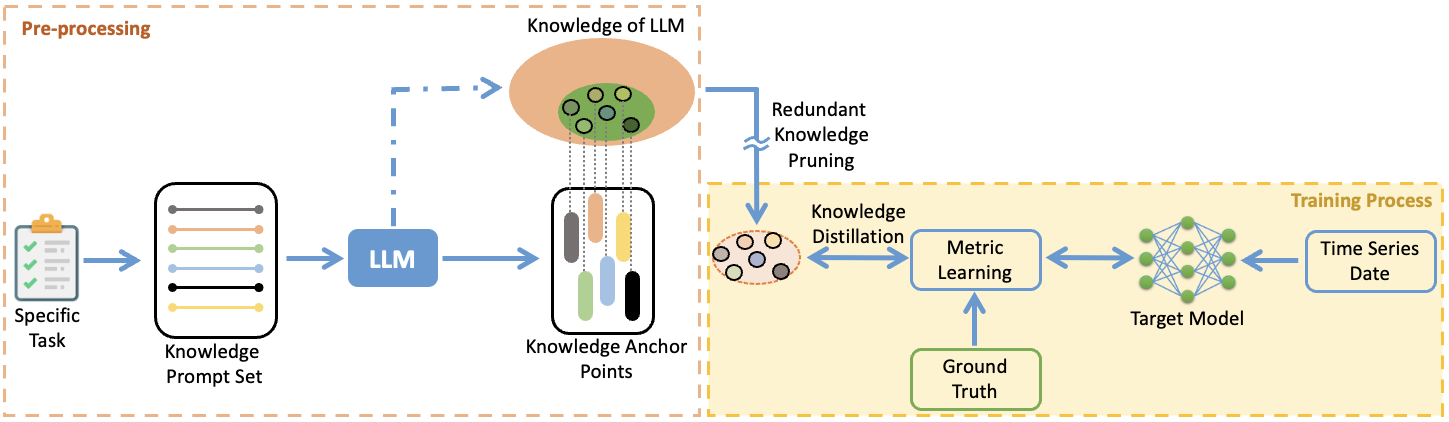}
	\end{center}
	\caption{The pipeline of our knowledge pruning (KP). Our KP coonsists of two stages: pre-processing and training stage. For a specific task, a knowledge prompt set is firstly produced, where these prompts are forwarded into a pre-trained LLM to obtain the corresponding language embeddings. Then, these embeddings are used as knowledge anchor points to estimate the pertinent knowledge and prune redundant knowledge of the LLM. After that, in the training stage, these knowledge anchor points are regarded as prior knowledge. The prior knowledges are transfered to the target mode via knowledge distillation.}
	\label{fig:pipeline}
\end{figure*}

Knowledge is implicitly stored in neural networks. It is difficult to directly obtain the specified network in general. To address this problem, we alleviate the dialogue scheme of LLMs to generate a series of language embeddings based on prompts. The pipeline of our KP is shown in Fig. \ref{fig:pipeline}, where our KP is composed of two stages: pre-processing and training stage. Given a specific downstream task, a knowledge prompt set (KPS) is firstly generated. After that, the prompts in KPS are forwarded into a pre-trained LLM to produce corresponding embeddings, which serve as knowledge anchor points (KAPs) and are used to represent the pertinent knowledge. We regard this pertinent knowledge as prior knowledge for the target model. After that, at the training stage, the metric learning and knowledge distillation are leveraged to transfer this prior knowledge into the target model. Additionally, the output based on metric learning is generally discrete. To extend the application of our KP to the tasks with continuous output, an anchor voting scheme (AVS) is proposed, which enables our KP to produce arbitrary value, achieving significant improvements on both classification and regression tasks.

\subsection{Knowledge Pruning}
Our knowledge pruning consists of three steps: knowledge prompt set generation, knowledge anchor point production and pertinent knowledge distillation.

\textbf{Knowledge Prompt Set Generation} Knowledge prompt set (KPS) contains the prompts which are forwarded into a pre-trained LLM for getting the knoweldge anchor points (KAPs). 
In this paper, these prompts in KPS indicate the description of corresponding data. Since we devise to apply our proposed KP to two fundumental tasks: regression and classification, two different prompt templates are proposed. In regression, the remaining sueful prediction is used to evaluate the performance of our KP, and the prompt template is ``\textit{The remaining useful life is \{num\}}.'', where \textit{num} indicates the correspoding groud truth value and ranges from [$y_{min}$, $y_{max}$] In classification, we apply our KP on the human activity recognition, and the prompt template is ``\textit{The subject is \{action\}}.'', where \textit{action} means the name of the corresponding activity.

\textbf{Knowledge Anchor Point Production} After obtaining the KPS, we forward these prompts in the KPS to a pre-trained LLM to obtain the language embeddings, which can be formulated as following.
\begin{equation}
	z_i = F_{l}(\mathcal{P}_{i}),
\end{equation}where $\mathcal{P}_{i}$ indicates the $i$th prompt, $F_l$ represents a pre-trained LLM, and $z_i$ is the produced language embedding termed as a knowledge anchor point (KAP). These KAPs are used to represent the space of pertinent knowledge.

\textbf{Pertinent Knowledge Distillation} Without transfering the entire knowledge of a LLM, our KP only transfer the pertinent knowledge which is indicated by KAPs. However, there is domain gap between the knowledge learned by LLM and the knowledge of downstream tasks. To alleviate this issue, an alignment module consisting of 2 fully connected layers are used to project these KAPs into tha latent space of the downstream task. This process is computed as below,
\begin{equation}
	k_i = \phi(z_{i}),
\end{equation}where $\phi$ means the alignment module and $k_i$ is the transformed feature vector, which serves as a prior knowledge.
Given a segment of time series data $x_i$, we forword it into our target model $F_t$, and get the predicted feature vector $f_i$. To utilize the prior knowledge $\mathcal{Z} = \{z_i | i = 1, \dots, N\}$, the metric learning is leveraged. 

Moreover, to optimze the target model and the alignment simultaneously, based on the unidirectional metric learning in Prototypical Networks \cite{snell2017prototypical}, we develope a bi-directional metric learning. For optimizing the target model, the process is computed as:
\begin{equation}
	p_t(i) = \frac{\exp(-d(k_i, x_i))}{\sum_{t=1}^{|\mathcal{Z}|}\exp(-d(k_t, x_i))}, 
\end{equation}where $d$ denotes the distance function. To improve the numerical stability and computational efficiency, we further improve this computation progress and compute the prediction as below:
\begin{equation}
	p_t(i) = \log(\frac{\exp({\rm simi}(k_i, f_i))}{\sum_{t=1}^{|\mathcal{Z}|}\exp({\rm simi}(k_t, f_i))}), 
\end{equation}where simi is the cosine similarity. For the alignment optimzation part, the process can be formulated as:
\begin{equation}
	p_l(i) = \log(\frac{\exp({\rm simi}(k_i, f_i))}{\sum_{t=1}^{|\mathcal{B}|}\exp({\rm simi}(k_i, f_t))}), 
\end{equation}where $|B|$ denotes the batch number.
Finally, to distill the pertinent knowledge to the target model, the Kullback–Leibler divergence (KL-div) is used and the final loss is:
\begin{equation}
	L = 0.5*D_{KL}(p_t, p_g) + 0.5*D_{KL}(p_l, p_g^{T}),
\end{equation}where $p_g$ is the ground truth distribution and is defined as following,
\begin{equation}
	p_g(i) = \frac{\exp(g_i * \tau)}{\sum_{t=1}^{|\mathcal{B}|}\exp(g_t*\tau))}, 
\end{equation}where $g_i$ is equal to one for correpsonding prompt description, while is zero. Similarly to knowledge distillation, $\tau$ is a temperature hyper-parameter.

\subsection{Anchor Voting Scheme}
Since our KP is based on metric learning, the prediction of the target model is discrete. To extend our KP to the task with continuous output like regression, an anchor voting scheme (AVS) is proposed.

Given the prediction distribution $\mathcal{S} = \{p_t(i)|i=1,\dots,|\mathcal{Z}|\}$, we firstly sort these scores in a descending order as below:
\begin{equation}
	\hat{\mathcal{S}} = {\rm sort}(\mathcal{S}).
\end{equation}After that, these scores are cumulated according to Eq. \ref{eq:cumsum}.
\begin{equation}\label{eq:cumsum}
	\mathcal{S}_a = {\rm cumsum}(\hat{\mathcal{S}})
\end{equation}Then, the cumlated scores which are larger than $\theta$, are formed into a voting set $\mathcal{V} = \{v_i| 1,\dots, |\mathcal{V}|\}$. The final prediciton is generated as following,
\begin{equation}
	o = \frac{\sum_{i=1}^{|\mathcal{V}|}v_i*n_i}{\sum_{i=1}^{|\mathcal{V}|}v_i},
\end{equation}where $n_i$ indicates the numerical value described by the KAP $v_i$. With our proposed  AVS, our proposed KP is effectively entended to the regression task, achieving significant performances.\cite{chen2020machine}

\section{Experiments}
\label{sec:exp}
To verify the effectiveness of our Knowledge Pruning (KP), extensive expriments are conducted in this section.

\begin{table*}[t!]
	\caption{Comparison with other methods in regression. Compared with other methods, our KP performs much better, achieving the best performances on nearly all subsets.}
	\begin{center}
		\scalebox{0.75}
		{\begin{tabular}{c|c|c|c|c|c|c|c|c|c|c}
				\hline
				Dataset & \multicolumn{2}{c|}{FD001} & \multicolumn{2}{c|}{FD002} & \multicolumn{2}{c|}{FD003} & \multicolumn{2}{c|}{FD004} & \multicolumn{2}{c}{AVG} \\
				\hline
				\hline
				Evaluation & RMSE & Score & RMSE & Score & RMSE & Score & RMSE & Score & RMSE & Score \\
				\hline
				Li et al. & 12.61 & 273.70 & {22.36} & {10412.00} & {12.64} & {284.10} & {23.31} & {12466.00} & 17.73 & 5858.95\\
				BLCNN & 13.18 & 302.27 & 19.09 & 1558.00 & 13.75 & 381.37 & 20.97 & 3859.00 & 16.75 & 1525.16 \\
				PE-Net & 13.98 & 280.87 & 14.69 & 881.73 & 12.33 & 272.85 & 15.40 & 1103.18 & 14.10 & 634.66 \\
				DGRU & 18.54 & 1467.00 & 20.06 & 4085.00 & 19.28 & 1488.00 & 20.88 & 3872.00 & 19.69 & 2728.00 \\
				AdaNet & 13.12 & 248.45 & 15.20 & 890.71 & 12.41 & 231.06 & 15.02 & 883.21 & 13.94 & 563.36 \\
				Jang et al. & 12.47 & 253.00 & {18.18} & {1618.00} & 11.88 & 270.00 & {22.11} & {2797.00} & 16.16 & 1234.5\\
				KDnet &13.68 & 362.08 & 14.47 & 929.20 & 12.95 & 327.27 & 15.96 & 1303.19 & 14.27 & 730.44 \\
				Two-Stream BiLSTM& \textbf{12.07} & 208.11 & 14.97 & 847.98 & 11.84 & 211.80 & 14.94 & 906.61 & 13.45 & 543.63\\
				\hline
				\hline
				KP (\textit{ours}) & 12.42  & \textbf{197.05} & \textbf{12.86} & \textbf{584.56} & \textbf{11.29} & \textbf{175.50} & \textbf{14.09} & \textbf{788.75} & \textbf{12.66} & \textbf{436.47}\\
				\hline	
		\end{tabular}}
	\end{center}
	\label{table:comp1}
\end{table*}
\subsection{Datasets and Experimental Setup}
\textbf{Datasets} To comprehensively investigate the performances of our KP, two fundamental tasks on edge-computing devices: classification and regression, are evaluated in this paper. In classificaiton, the human activity recognition (HAR) task is studied and four different benchmarks: UCI\_HAR \cite{anguita2013public}, Opportunity \cite{roggen2010collecting}, PAMAP2 \cite{reiss2012introducing}, and WISDM \cite{kwapisz2011activity} are used. These benchmarks contain different number of activity categories ranging from 6 to 17 with different scales between 3k and 29k samples. In regression, the remainning useful life (RUL) prediciton is alleviated for evaluation, where the C-MAPSS \cite{saxena2008damage} dataset is used. C-MAPSS contains four different subsets: FD001, FD002, FD003 and FD004 with different scenarioes. 

\textbf{Experimental Setup} In classifiction, for consistency and meaning fulcomparison, the training and inference process on UCI\_HAR, Opportunity and  PAMAP2 are conducted according to the protocol of iSPLInception \cite{ronald2021isplinception}. Since the experiments in iSPLInception \cite{ronald2021isplinception} do not include WISDM benchmark,  the expriments on WISDM follow the setting in Multi CNN-BiLSTM \cite{challa2022multibranch}. For fair comparion, other compared methods are re-implemented under the same setting. According to approches \cite{ronald2021isplinception,challa2022multibranch}, F1-Score is used for evaluation in HAR tasks. 
In regression, some methods are also re-implemented under the same conditions. The training and inference processes are conducted according to classic RUL methods \cite{jin2022bi,chen2020machine}. RMSE and scoring fuction are used as evaluation metrics. Two hyper-parameters $\tau$ and $\theta$ are set as 10 and 0.9, respectively for all experiments. The pre-trained text encoder in CLIP \cite{radford2021learning} is used as the pre-trained LLM in experiments. Experiments are conducted on a workstation with a GeForce RTX 4080 GPU and 128 GB memory from 1 to 4 hours. 

\subsection{Comparison with other methods}
To evaluate the performances of our KP, we compare our approach with orther state-of-the-art (SOTA) methods. The exprimental results in regression and classificationare listed in Table. \ref{table:comp1} and Table. \ref{table:comp2}, respectively.

In the RUL task, serveral SOTA approaches: Li et al. \cite{li2018remaining}, BLCNN \cite{liu2019novel}, PE-Net \cite{jin2022position}, DGRU \cite{behera2021generative}, AdaNet \cite{jin2023adaptive}, Jang et al. \cite{jang2021siamese} and KDnet \cite{xu2021kdnet} , are compared with our method. Among these methods, Li et al. proposes a CNN based network to predict the RUL. PE-Net integrates position encoding scheme with an optimzed CNN architecture for the RUL task. AdaNet introduces the deformable convolution into the RUL task. BLCNN devises a hybrid network which combines RNN and CNN together to improve the prediction accuracy. DGRU apply the adversarial learning on the RUL task. A self-supervised learning approch is proposed in Jang et al. KDnet utilize knowledge distillation to transfer the knowledge in RNN to a CNN model. Benefitted from the learned pertinent knowledge from a pre-trained LLM, our KP performs much better than them and achieve the best performances.

\begin{table}[htbp]
	\caption{Comparison with other methods in classification. Compared with other methods, our proposed KP achieves the best performances in F1-Score among four different HAR benchmarks.}
	\begin{center}
		\scalebox{0.9}
		{\begin{tabular}{c|c|c|c|c}
				\hline
				Methods & UCI\_HAR & Oppotunity & PAMAP2 & WISDM \\
				\hline
				LSTM-CNN & 93.14 & 78.19 &72.22	&96.02\\
				CNN & 93.21 & 79.73 &62.29	&95.51\\
				Multi CNN-GRU & 94.05 & 83.92& 68.13&	96.15\\
				Multi CNN-BiLSTM & 93.60 & 84.53& 70.23&	95.8\\
				GRU\_INC & 93.67 & 72.58 &75.88&	83.93 \\
				DTL & 93.11 & 74.03& 82.16&	96.42\\
				iSPLInception & 93.08 & 84.45 &80.01&96.14\\
				\hline
				KP (\textit{ours}) & \textbf{96.63} & \textbf{86.74} & \textbf{85.28} & \textbf{98.25} \\
				\hline	
		\end{tabular}}
	\end{center}
	\label{table:comp2}
\end{table}

In the HAR task, we compare our KP with seven SOTA methods: LSTM-CNN \cite{xia2020lstm}, CNN \cite{van2020mcfly}, Multi CNN-GRU \cite{dua2021multi}, Multi CNN-BiLSTM \cite{challa2022multibranch}, GRU\_INC \cite{mim2023gru}, DTL \cite{ige2023deep} and iSPLInception \cite{ronald2021isplinception}. These compared approaches employ different network architectures like RNN, CNN and even hybrid networks. As a novel comdel compression paradigm, our proposed KP is fundamentally orthogonal to existing HAR methods and can be applied to any existing HAR approaches. We apply our KP to two different methods: DTL and iSPLInception, and list the best performances we achieved in Table \ref{table:comp1}. Through experiments, it demonstrates that our KP effectively transfer the pertinent knowledge of a pre-trained LLM to the target model, which achieves the best performances among other SOTA methods.

\begin{table*}[t!]
	\caption{Ablation study in regression. Baseline1 indicates the Bi-LSTM, baseline2 is the PE-Net, and baseline3 represents the Two-Stream BiLSTM. With our proposed KP, the performances on three different baselines are improved by a large marge.}
	\begin{center}
		\scalebox{0.78}
		{\begin{tabular}{c|c|c|c|c|c|c|c|c|c|c}
			\hline
			Dataset & \multicolumn{2}{c|}{FD001} & \multicolumn{2}{c|}{FD002} & \multicolumn{2}{c|}{FD003} & \multicolumn{2}{c|}{FD004} & \multicolumn{2}{c}{AVG} \\
			\hline
			\hline
			Evaluation & RMSE & Score & RMSE & Score & RMSE & Score & RMSE & Score & RMSE & Score \\
			\hline
			Baseline1 & 13.09 & 260.67 & 15.85 & 871.62 & 13.23 & 266.12 & 15.81 & 1065.47 & 14.50 & 615.97 \\
			Baseline1+KP (\textit{ours}) & 12.82 & 201.85 & 14.09 & 716.48 & 11.59 & 190.94 & 15.17 & 953.94 & 13.42 & 515.80 \\
			\hline
			Baseline2 & 13.98 & 280.87 & 14.69 & 881.73 & 12.33 & 272.85 & 15.40 & 1103.18 & 14.10 & 634.66 \\
			Baseline2 + KP(\textit{ours}) &13.63 & 251.64 & 14.11 & 721.38 & 12.44 & 197.96 & 15.52 & 938.36 & 13.92 & 527.34 \\
			\hline
			Baseline3 & \textbf{12.07} & 208.11 & 14.97 & 847.98 & 11.84 & 211.80 & 14.94 & 906.61 & 13.45 & 543.63\\
			Baseline3 + KP (\textit{ours}) & 12.42  & \textbf{197.05} & \textbf{12.86} & \textbf{584.56} & \textbf{11.29} & \textbf{175.50} & \textbf{14.09} & \textbf{788.75} & \textbf{12.66} & \textbf{436.47}\\
			\hline	
		\end{tabular}}
	\end{center}
	\label{table:abla1}
\end{table*}

\begin{table*}[t!]
	\caption{Ablation study for AVS. Baseline1 indicates the Bi-LSTM. Through experiments, it shows that without our proposed AVS, the performances of KP on the regression task are limited. After applying our AVS, the performances on the regression task are significantly improved.}
	\begin{center}
		\scalebox{0.74}
		{\begin{tabular}{c|c|c|c|c|c|c|c|c|c|c}
				\hline
				Dataset & \multicolumn{2}{c|}{FD001} & \multicolumn{2}{c|}{FD002} & \multicolumn{2}{c|}{FD003} & \multicolumn{2}{c|}{FD004} & \multicolumn{2}{c}{AVG} \\
				\hline
				\hline
				Evaluation & RMSE & Score & RMSE & Score & RMSE & Score & RMSE & Score & RMSE & Score \\
				\hline
				Baseline & 13.09 & 260.67 & 15.85 & 871.62 & 13.23 & 266.12 & 15.81 & 1065.47 & 14.50 & 615.97 \\
				Baseline+KP w/o AVS (\textit{ours}) & 16.07 & 339.03 & 14.27 & 963.37
				& 13.35 & 219.47 & 17.29 & 1340.89  & 15.25 & 715.69 \\
				Baseline+KP (\textit{ours}) & \textbf{12.82} & \textbf{201.85 }& \textbf{14.09} & \textbf{716.48} & \textbf{11.59 }& \textbf{190.94} & \textbf{15.17} & \textbf{953.94} & \textbf{13.42} & \textbf{515.80} \\
				\hline	
		\end{tabular}}
	\end{center}
	\label{table:avs}
\end{table*}

\subsection{Ablation Study}
To verify the effectiveness, experiments on ablation study are presented. Our KP is orthogonal to approaches for time series analytics and can be directly applied to these methods. To show the generalization of our KP, we apply our KP on serveral different networks and show the performance improvements. Experimental results on the regression task, RUL, and classification task, HAR are listed in Table. \ref{table:abla1} and Table. \ref{table:abla2}, respectively. 

In the RUL task, three different approaches: Bi-LSTM, Two-Stream BiLSTM \cite{jin2022bi} and PE-Net \cite{jin2022position} are used as our baselines. Bi-LSTM is a shallow network, which consists of two bi-directional LSTM. Two-Stream BiLSTM integrates the handcrafted feature flow  \cite{jin2022bi}  into the raw time series data via a Bi-LSTM based network. PE-Net designs a CNN with a position encoding scheme to predict the RUL. In Table. \ref{table:abla2}, Bi-LSTM is used as baseline1, PE-Net  is used as baseline2 and Two-Stream Bi-LSTM is used as baseline3. With our proosed KP, all these three methods are remarkablely improved among four different scenarioes. Compared with RMSE, Score is generally regarded as a more important evaluation metric, since it give more penalty on the late prediciton, which is similar to the practical setting. Among these three baselines, baseline3 acheves the best performances on average. After applying our KP, its performances are further improved by 19.7\% in Score.

In the HAR task, two different methods: DTL \cite{ige2023deep} and iSPLInception \cite{ronald2021isplinception}, are used as our baselines. DTL is a hybrid network, which combines CNN and RNN togethor to capture the temporal features. In comparison, iSPLInception proposes to utilize the inception based CNN network to classify the human activaities.In Table. \ref{table:abla1}, baseline1 indicates the DTL, and basleine2 represents the iSPLInception. According to the experimental results, our KP is able to effectively improve the performances on these two baselines among all four benchmarks. The improvements on these methods range from 0.8\% to 13.7\%.

\begin{table}[htbp]
	\caption{Ablation study in classification. Baseline1 indicates the DTL method, and baseline2 represents the iSPLInception method. Our proposed KP is able to effectively improve the performances on two different network architectures among four different HAR benchmarks.}
	\begin{center}
		\scalebox{1.0}
		{\begin{tabular}{c|c|c|c|c}
				\hline
				Methods & UCI\_HAR & Oppotunity & PAMAP2 & WISDM \\
				\hline
				Baseline1 & 93.11 & 74.03& 82.16&	96.42\\
				Baseline1+KP (\textit{ours})& \textbf{96.63} & 84.14 & \textbf{85.28} & 97.18\\
				\hline
				Baseline2) & 93.08 & 84.45 &80.01&96.14\\
				Baseline2 + KP (\textit{ours}) & 94.75 & \textbf{86.47} & 83.99 & \textbf{98.25}\\
				\hline	
		\end{tabular}}
	\end{center}
	\label{table:abla2}
\end{table}

These expriments above show that our propsoed KP effectively identify the pertinent knowledge and transfer it to the target model. With our KP, all these five baselines are improved by a large marge.

\textbf{Effectiveness of AVS} AVS is proposed to enable the metric learning based network to predict continuous value for the regression task, like RUL. Experiments are designed to show the effectiveness of our proposed AVS, which are listed in Table \ref{table:avs}. The Bi-LSTM is used as the baseline. The experimental reults indicates that without our propsoed AVS, the performances of KP on the regrression task are not satisfactory. After applying our proposed AVS, our KP can effectively improve the accuracy on the RUL prediction by a large marge.

Based on the experiments above, it can be found that our proposed KP can consistently improve the performances across different tasks and benchmarks. Since the improvemment by KP ranges from 0.8\% to 13.7\%, the effectiveness of our KP may be affected by the specific data distribution and the neural network architecture.

\subsection{Computation Efficiency}

Our KP is proposed to alleviate the issue of the computation cost in LLMs. The experiments on computation efficiency is carries out and listed in Table. \ref{table:comput}, where the FLOPs and Params are listed to compare the computation efficiency.

\begin{table}[htbp]
	\caption{Experiments on computation efficiency.}
	\begin{center}
		\scalebox{.9}
		{\begin{tabular}{c|c|c}
				\hline
				Methods & FLOPs (G) & Params (M) \\
				\hline
				LLM in CLIP & 5.96 & 63.43 \\
				Two-Stream BiLSTM + KP (\textit{ours}) & 0.002 & 0.042 \\
				DTL+ KP (\textit{ours}) & 0.024 & 1.12 \\
				\hline	
		\end{tabular}}
	\end{center}
	\label{table:comput}
\end{table}
As listed in Table. \ref{table:comput}, we apply our KP to two networks: Two-Stream BiLSTM  and DTL for the task RUL and HAR, respectively. Since DTL applies a hybrid network, which is composed of CNN and RNN and is more complex than the Two-Stream BiLSTM, the computation complexity of DTL is higher than that of Two-Stream BiLSTM. Nevertheless, the computation demands on these two approaches are much lower than that of the LLM in CLIP. According to the experimental results, our proposed KP is able to effectively prune the redundant knowedlge of LLM. The computation issue of LLMs is well alleviated, and the performances of the target model are improved.

\begin{figure}[htbp]
	\begin{center}
		\includegraphics[width=0.65\linewidth]{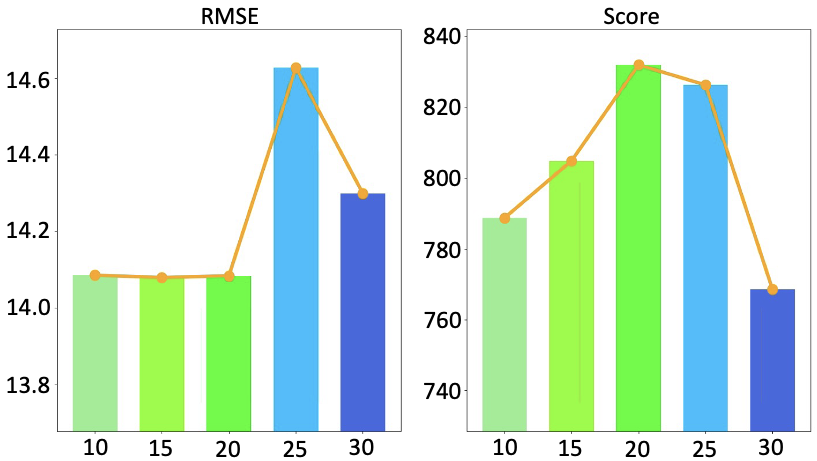}
	\end{center}
	\caption{Expriements on RUL task with different $\tau$ values on FD004.}
	\label{fig:exp_rul}
\end{figure}

\subsection{Sensitivity Analysis}

\begin{figure}[htbp]
	\begin{center}
		\includegraphics[width=.65\linewidth]{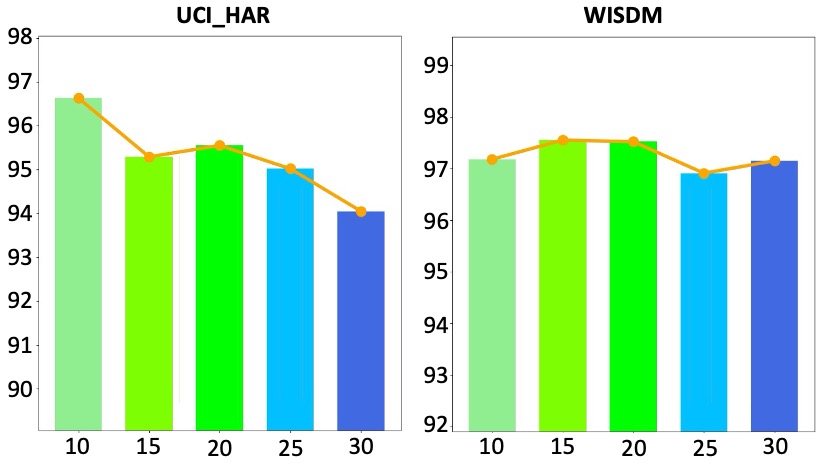}
	\end{center}
	\caption{Expriements on HAR task with different $\tau$ values.}
	\label{fig:exp_har}
\end{figure}

Our proposed KP involves two hyper-parameters: $\tau$ and $\theta$. To investigate the impact of different values of these two parameters, several experiments are conducted and discussed. For the parameter $\tau$, we graduately increase its values and carry out experiments on HAR and RUL tasks, respectively. These experimental results on RUL and HAR are illurstrated in Fig. \ref{fig:exp_rul} and Fig. \ref{fig:exp_har}, respectively.

In Fig. \ref{fig:exp_rul}, our KP is applied on Two-Stream BiLSTM with different $\tau$ values. Experiments are carried out in FD004 subset, which contains the most complex scenarioes. Although the difference performances are obtained on RUL task, their performances are still better than the baseline. With our KP, the performances of Two-Stream BiLSTM are consistly improved under different values of $\tau$.

In Fig. \ref{fig:exp_har}, we apply our KP on the DTL in two different datasets: UCI\_HAR and WISDM benchmarks. It shows that our KP can improves the performances of DTL under different $\tau$ values.

AVS is proposed to regression tasks, which enables our network to predict continuous value for the RUL task. The hyper-parameter $\theta$ in AVS is used as a threshold value to select the anchor for voting. To investigate the stability of our AVS, we apply our KP to the Two-Stream BiLSTM  and  design expriments on FD004 with different $\theta$, which are presented in Table. \ref{table:theta}.

\begin{table}[htbp]
	\caption{Experiments for AVS with different $\theta$ value on FD004.}
	\begin{center}
		\scalebox{.9}
		{\begin{tabular}{c|c|c|c|c|c|c}
				\hline
				Metric & baseline&0.9 & 0.8 & 0.7  & 0.6  & 0.5  \\
				\hline
				RMSE & 14.94 &14.09 & 13.92 & \textbf{13.91} & 14.11 & 14.30 \\
				Score & 906.61 &\textbf{788.75} & 810.19 & 824.61 & 853.73 & 891.58\\
				\hline	
		\end{tabular}}
	\end{center}
	\label{table:theta}
\end{table}

As listed in Table \ref{table:theta}, our AVS with different $\theta$ values is able to consistly improve the performances of Two-Stream BiLSTM. This demonstrates that our proposed AVS is adaptive to the variation of $\theta$.

\section{Conclusions}
\label{sec:conslusion}
In this paper, we have proposed a new model compression paradigm, Knowledge Pruning (KP). Our KP consists of three steps: knowledge prompt set generation, knowledge anchor point production and pertinent knowledge distillation. Furethurmore, since our KP is based on metric learning, the performances on the regresison tasks may be limited. To extend our KP to the regression task, a anchor voting scheme has been proposed. Through experiments, our KP has effectively pruned the redundant knoweldge of LLMs for a specific downstream task and accurately transfer the pertinent knowledge to the target model. With our KP, the computation cost introduced by LLMs is largely reduced, and satisfacotry performances are achieved. Our KP shown siginificant improvement on both classification task, HAR and regression task, RUL, achieving state-of-the-art performances.

\bibliographystyle{unsrtnat}
\bibliography{ref}

\end{document}